\documentclass[10pt,twocolumn,letterpaper]{article}

\usepackage{cvpr}

\usepackage{graphicx}
\usepackage{amsmath}
\usepackage{amssymb}
\usepackage{booktabs}
\usepackage{color}
\usepackage{multirow}
\usepackage{float}

\usepackage[pagebackref,breaklinks,colorlinks]{hyperref}

\usepackage[capitalize]{cleveref}
\crefname{section}{Sec.}{Secs.}
\Crefname{section}{Section}{Sections}
\Crefname{table}{Table}{Tables}
\crefname{table}{Tab.}{Tabs.}

\def\cvprPaperID{*****} 
\def\confName{CVPR}
\def\confYear{2025}

\begin{document}

\title{Anatomically-Grounded Fact Checking of Automated Chest X-ray Reports}

\author{R. Mahmood, D. M. Reyes, G. Wang, P. Yan\\
Rensselaer Polytechnic Institute\\
110 8th St, Troy, NY 12180\\
{\tt\small mahmor@rpi.edu}
\and
P. Kaviani, M. Kalra, \\
Massachusetts General Hospital\\
Harvard University, USA\\
{\tt\small kaviani@mgh.harvard.edu}
\and
K.C.L. Wong, N. D'Souza, L. Shi, J. Wu, T. Syeda-Mahmood, \\
IBM Research - Almaden\\
650 Harry Road, San Jose, USA\\
{\tt\small stf@us.ibm.com}
}
\maketitle

\vspace{-0.2in}
\begin{abstract}
\noindent With the emergence of large-scale vision-language models, realistic radiology reports may be generated using only medical images as input guided by simple prompts. However, their practical utility has been limited due to the factual errors in their description of findings. In this paper, we propose a novel model for explainable fact-checking that identifies errors in findings and their locations indicated through the reports. Specifically, we analyze the types of errors made by automated reporting methods and derive a new synthetic dataset of images paired with real and fake descriptions of findings and their locations from a ground truth dataset. A new multi-label cross-modal contrastive regression network is then trained on this datsaset. We evaluate the resulting fact-checking model and its utility in correcting reports generated by several SOTA automated reporting tools on a variety of benchmark datasets with results pointing to over 40\% improvement in report quality through such error detection and correction. 
\end{abstract}

\section{Introduction}
\label{sec:intro}
\begin{figure}
\centering
  \centerline{\includegraphics[width=3.4in]{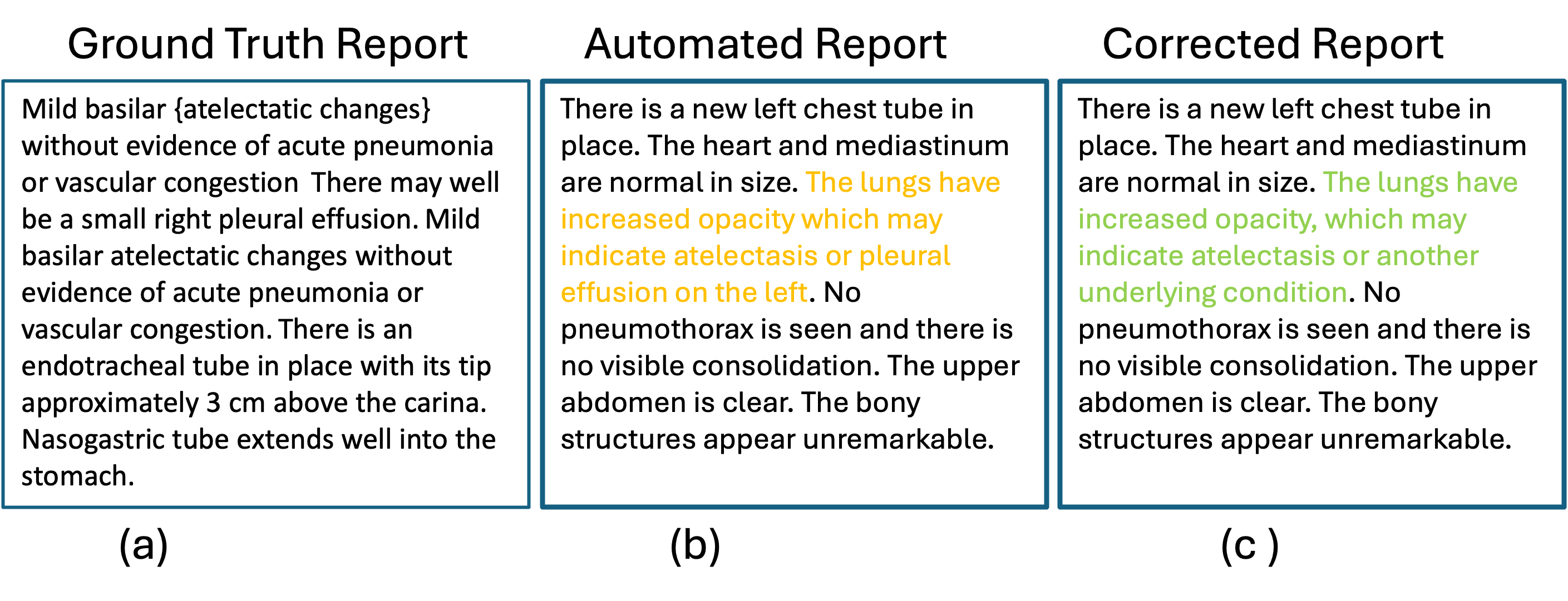}}
  \caption{Illustration of errors in radiology reporting. (a) Ground truth report. (b) Generated report by XrayGPT\cite{xraygpt}. (c) Corrected report by our method. The sentence with error in finding is colored orange in (b) and corrected sentence is shown in green in (c). }\medskip
  \label{errorillust}
\end{figure}

With the emergence of large-scale vision-language models (VLMs), several researchers have turned to medical applications of automated report generation for medical images such as chest X-rays~\cite{maira,Pang2023,syeda-mahmood2020,Gao2024,Ranjit2023,Ramesh2022,Nguyen2021,Endo2021-2}. A preliminary radiology report produced by such models is helpful in emergency room settings where radiologists may not be readily available and the interpretation needs to be performed by residents or other clinical staff. However, the predominance of hallucinations and factual errors have made such report generators less practical in clinical workflows. Figure~\ref{errorillust}b shown an example of the such an error in an automatically generated report in the sentence highlighted.  The corresponding ground truth report fragment is shown in Figure~\ref{errorillust}a.  

Methods for detecting and correcting hallucinations in large language models (LLMs, VLMs) have primarily been developed for use during training or fine-tuning \cite{Passi2022,nieman,midas,Li2023,Gunjal2023,yin2023woodpecker,zhou2023analyzing}. Fact-checking methods that are available during inference often consult external textual resources such as Wikipedia or do a general assessment based on linguistic cues. These are also not suitable for radiology reports which contain clinically specific descriptions of the associated medical image.  Recently, a simple SVM classifier was developed for fact-checking that leverages radiology images but was trained to identify and correct a single finding error in specific radiology report sentences,  making them less generally applicable to handle a wide class of factual errors made by modern automated reporting tools\cite{mlmi2023}. Thus while there is a large body of work on radiology report generation, there is a paucity of fact-checking methods for radiology report correction. 
\begin{figure}
\centering
  \centerline{\includegraphics[width=3.5in]{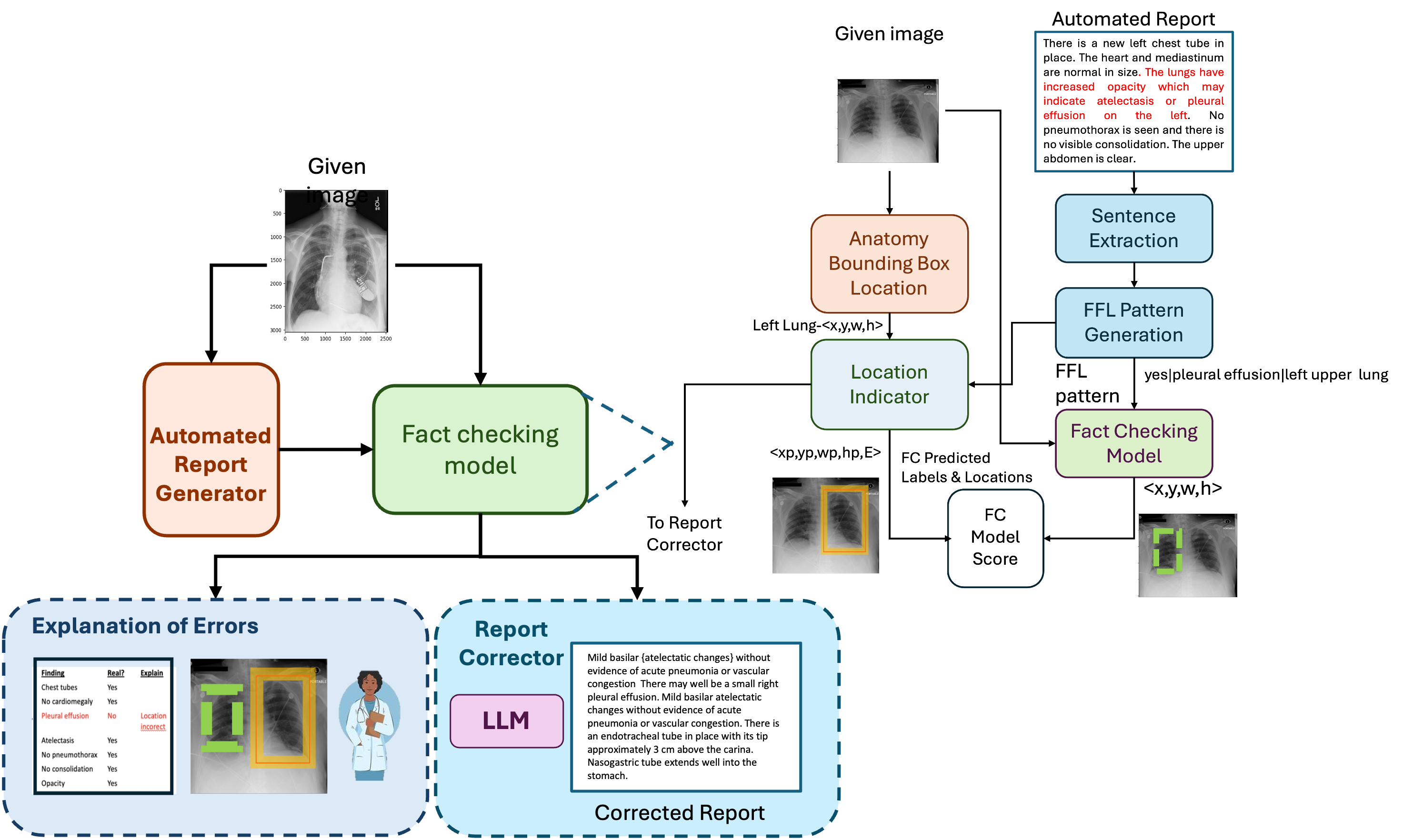}}
  \caption{Illustration of the a fact-checking system for clinical workflows. An automatically generated report is evaluated by the fact checking (FC) model and an explanation generated documenting the finding errors and their localization issues. A report corrector LLM then uses the fact-checking results and the original report to produce a corrected report.}
  \label{radiologyworkflow}
\end{figure}
\begin{figure*}
\centering
(a)\hspace{1in}(b)\hspace{1in}(c)\hspace{1in}(d)\hspace{1in}(e)
  \centerline{\includegraphics[width=6.5in]{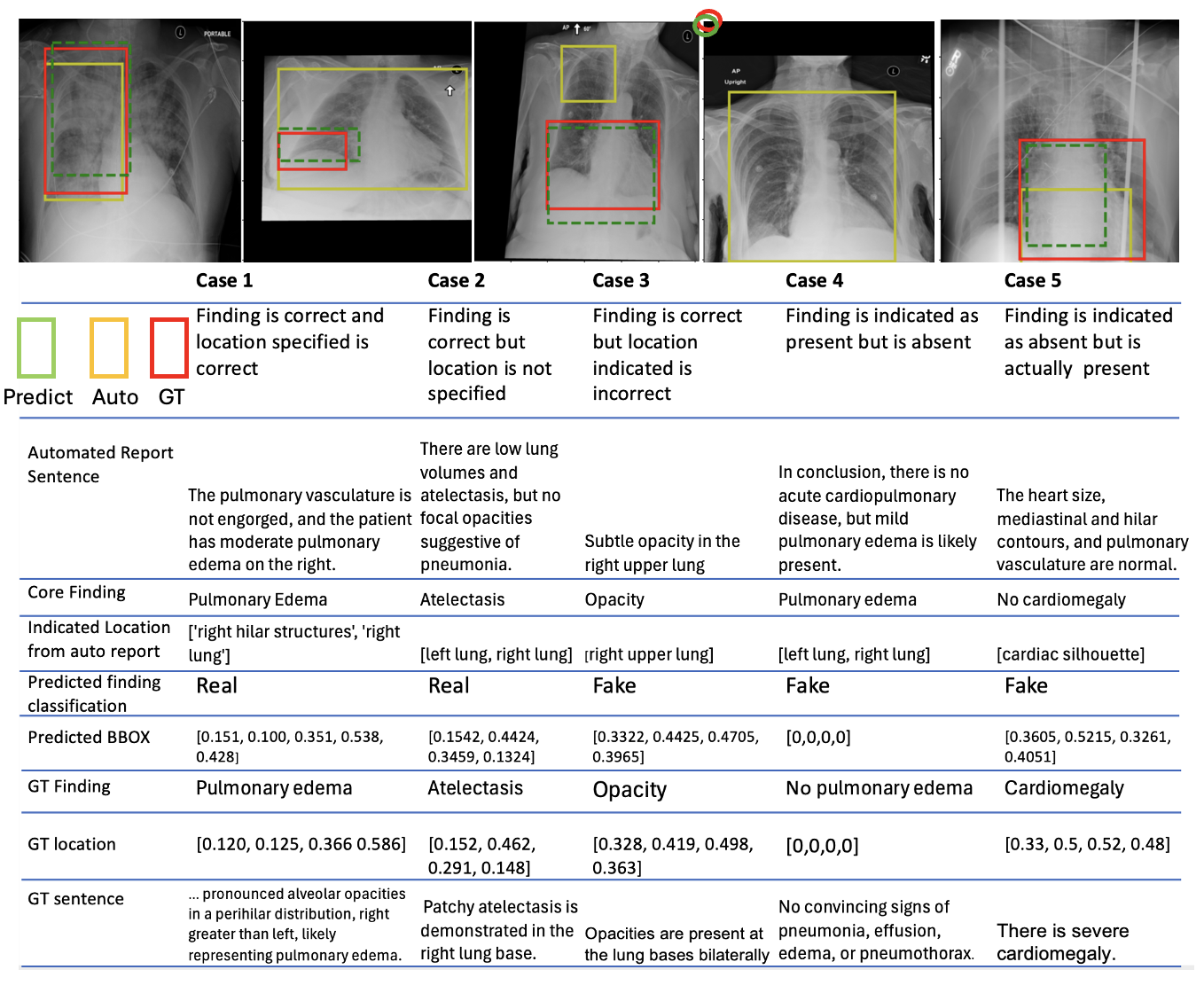}}
  
  \caption{Illustration of fact-checking on automatically generated reports. 5 cases are shown including a case of no error flagged as  real finding by our FC model. For cases on absence finding (e.g. case 4), the predicted and ground truth location is at <0,0> coordinate as explained in text. The predicted finding location is in Green, while the ground truth location in red and the indicated location from automated report in yellow/orange. } 
  \label{exampleresults}
\end{figure*}


In this paper, we introduce an innovative method of explainable fact-checking and correction for automatically produced chest x-ray radiology reports. Specifically, we analyze the types of errors made by automated reporting methods to develop a novel synthetic dataset of images paired with real and fake findings, which are obtained through perturbation of their identities and location descriptions from ground truth reports. We then develop a new  multi-label contrastive regression network for fact-checking that chains a multi-label supervised contrastively-learned encoder with a regression classifier to classify and anatomically ground the findings. The associated sentences in the reports containing the incorrect findings are then edited and reformed into valid sentences through a large language-model to produce the corrected reports. Figure~\ref{radiologyworkflow} depicts our overall approach to fact-checking. 

Results of testing on multiple X-ray datasets demonstrate the robustness of the method in terms of accuracy of prediction, and localization.  We also evaluate its utility in correcting automated reports generated by several SOTA automated reporting tools on a variety of benchmark datasets to lead to 40\% improvement in the quality of automated reports. 

Our paper makes the following novel contributions:
\begin{itemize}
\item We propose for the first time, an anatomically-grounded fact-checking model (FC model) to help create a surrogate ground truth during inference in clinical workflows. The fact-checking model proposed itself is a novel multi-label cross-modal contrastive regression network. 
\item We develop a novel synthetic dataset of image-finding description pairs capturing the range of errors made by automated reports generators for chest x-ray images. This will be contributed to open source. 
\item To our knowledge, we are also the first to use a large language model for fact-checking guided radiology report correction which when combined with FC model leads to over 40\% improvement in report quality.  
\end{itemize}

\section{Related Work}

Current methods for detecting and correcting for errors in generative AI reports have been primarily developed for training and fine-tuning large language models or vision-language models\cite{Passi2022,nieman,midas,Li2023,Gunjal2023,yin2023woodpecker,zhou2023analyzing}. They use direct policy optimization (DPO)\cite{rafailov2023direct} or proximal policy optimization (PPO) \cite{zheng2023secrets} along with reward models\cite{ziegler2019finetuning} to assess fine-grained subjective performance using the reinforcement learning with human feedback(RLHF) paradigm.
While capturing human feedback is possible through Mechanical Turks for general LLM or VLMs, building similar reward models would be difficult for radiology reports needing large clinician time and attention. 
Methods for fact-checking during inference exist primarily for language-only models where patterns of phrases found repeatedly in text are used to spot errors or by consulting other external textual sources for checking the veracity of information in an agentic fashion\cite{Passi2022,nieman,midas, Thorne2018}. 
More recently, language models are also being used for fact-checking other LLM-generated reports \cite{Schmidt2024} which are not as suitable for radiology reports since these models themselves have errors.  Bootstrapping them with an independent source of verification would still be desirable. 

The closest work to us is an image-driven fact checking method described in \cite{mlmi2023} which reused a  vision-language model (CLIP) pre-trained on chest X-ray data and a binary SVM classifier to classify findings as real or fake in automated radiology reports\cite{mlmi2023}.  This approach, while promising had several limitations. First, since radiology report sentences describe multiple findings in a sentence, using such full sentences for training the model could make an entire sentence misclassified and removed during report correction.  Using full-length sentences also binds this method to sentence styles used in the training data and limits its generalizability across report generators. Next, the CLIP contrastive model used was pre-trained only on real pairs of images and reports and so features derived from such models are not suitable for discrimination between real and fake findings. Finally, the binary classifier used does not offer explanation of the errors nor anatomically locate the finding as done in our approach.  

While the general problem of correcting generated text from language models has been studied during training, inference, and post-hoc phases\cite{Yu2023, Schmidt2024, Pan2024}, correction for radiology reports so far have been through simplistic methods in which the entire sentence containing the finding is removed\cite{mlmi2023}. Further, report evaluation methods exist for assessing automated reports but need ground truth reports for comparison limiting their use at inference time in clinical workflows\cite{bleuscore,Ziegelmayer2023,rouge,bertscore,radgraph,Liu2023quality,Yu2023}.

\section{Extracting findings and their locations}
\label{prepare}
To make our fact-checking approach agnostic to automated reporting tools' sentence writing styles, we need to abstract the findings described in reports into structured representations.  We adopt the fine-grained finding patterns (FFL) work described in \cite{syeda-mahmood2020}, and restrict them to cover the core finding and its anatomical location as:
\begin{equation}
    F_{i}= <T_{i}|N_{i}|C_{i}|A_{i}>
    \label{ffleqn}
\end{equation}
\noindent where $T_{i}$ is the finding type, $N_{i}= yes|no$ indicates a present or absent finding respectively, $C_{i}$ is the normalized core finding name, $A_{i}$ is the anatomical location specified with laterality.  
Each finding is normalized to a standard vocabulary (e.g. enlarged cardiac silhouette versus cardiomegaly) using a  comprehensive clinician-curated chest X-ray lexicon reported in \cite{wu2020a,syeda-mahmood2020a}. Based on the vocabulary captured in the lexicon, a total of 101,088 distinct FFL patterns can be formed which are sufficient to capture the variety of findings reported in automated reporting tools. The FFL label extraction algorithm reported in \cite{syeda-mahmood2020} is known to be highly accurate in terms of the coverage of findings with around 3\% error mostly due to negation sense detection. More details can be found in \cite{syeda-mahmood2020}.

In addition to finding descriptions, we also use the anatomical location algorithm described in \cite{Wu2020-isbi2020,Wu2021} to locate bounding boxes in any frontal chest x-ray image for the 36 anatomical regions cataloged in the chest x-ray lexicon \cite{wu2020a,syeda-mahmood2020a}. The localization accuracy of the bounding box detector was previously assessed at 0.896 precision and 0.881 recall and was used to reliably generate the ChestImagenome benchmark dataset\cite{Wu2021}. 
\begin{table}
  \centering
  \begin{tabular}{l|l|l}
    \toprule
    Synthetic  & Generated  & Label ($<$xy,w,h,E$>$) \\
    Perturbation & Finding&\\
    \hline
   Original 	& yes$|$edema 	& $<0.14, 0.13, 0.72, 0.56, 1>$\\
Reversal 	& no$|$edema 	& $<0,0,0,0,0>$\\
Relocate 	& yes$|$edema 	& $<0.85,0.74, 0.10, 0.21, 0>$\\
Relocate 	& yes$|$edema 	& $<0.90,0.70,0.10,0.20,0>$\\
Substitution 	& yes$|$lung cyst 	& $<0.02,0.48, 0.10, 0.14, 0>$\\
    \bottomrule
  \end{tabular}
  \caption{Illustration of synthetic perturbations to produce the labeled dataset for training the FC model. For simplicity, we show only the core finding in column 2. }
  \label{synthetic}
\end{table}
\subsection{Developing a synthetic dataset}
\label{syntheticsection}
Given a dataset of chest X-rays and their associated reports, we extract all real FFL patterns and anatomical locations of regions covered by the FFL patterns. We then derive a synthetic dataset starting from these real FFL patterns to reflect the types of errors made by automated reporting tools. As reported in \cite{Yu2023}, these errors include false predictions, omissions, incorrect finding locations or incorrect severity assessments. In this paper, we focus on modeling incorrect findings and their locations.  

The synthetic dataset created for training our fact-checking model can be described in terms of finding-location (FL) pairs. Let F be the total list of possible findings in chest X-ray datasets.  Let $<I,R>$ be the sample set of corresponding image-report pairs in a gold dataset $D$. Since each report $R_{i}$ will contain a variable number of findings, an existing multi-label set of sample $D_{i}\in D=<I_{i},R_{i}>$ can be denoted by its real FL pairs  $FL_{iReal}=\{fl_{ij}\}=\{<f_{ij},l_{ij}>\}$ where:
\begin{equation}
    f_{ij} =<T_{ij}|N_{ij}|C_{ij}>,
    l_{ij}=<x_{ij},y_{ij},w_{ij},h_{ij}>
\end{equation}
Here $f_{ij}\in F_{iReal} $ is the jth real finding in report $R_{i}$  and $l_{ij}$ is the bounding box for the finding $f_{ij}$ in image $I_{i}$ starting at $(x_{ij},y_{ij})$ of width $w_{ij}$ and height $h_{ij}$ in normalized coordinates ranging from 0 to 1. 
Since locations are being modeled through $l_{ij}$, we drop the textual description $A_{j}$ from the FFL pattern $f_{ij}$ for purposes of FC model generation while retaining it still for report evaluation. 

Let $L_{j}=\{l_{ij}\}$ be the list of all normalized locations accumulated across all images of $D$ for a finding $F_{j}$. Randomly drawing from this set ensures that a synthetic location generated for $F_{j}$ is a valid location for some image in the dataset. Given a real finding $f_{ij}$ at location $l_{ij}$ for a sample $D_{i}$, we derive a set of fake finding-location pairs to simulate the potential errors. Specifically we create 3 variants to reflect (a) reversal of polarity  (b) relocation of the finding (c) substitution with and  without relocation as given below:
\begin{equation}
  FL_{iFake}=\{<\overline{fl_{ij}}, fl_{ik},fl_{mn}> \}
\end{equation}
where $\overline{fl_{ij}}$ is the reversed finding, $fl_{ik}$ is finding $f_{ij}$ relocated to a random new position $l_{k}\in L_{j}$, and $fl_{mj}$ is obtained by randomly substituting finding $f_{j}$ with $f_{m}\not\in F_{i}$ at location $l_{n}\in L_{m}$. 

Table~\ref{synthetic} shows synthetic perturbations created from an original finding "yes$|$edema" based on the operations above. 

\section{Developing a fact-checking model}
\begin{figure}
\centering
 \centerline{\includegraphics[width=2.5in]{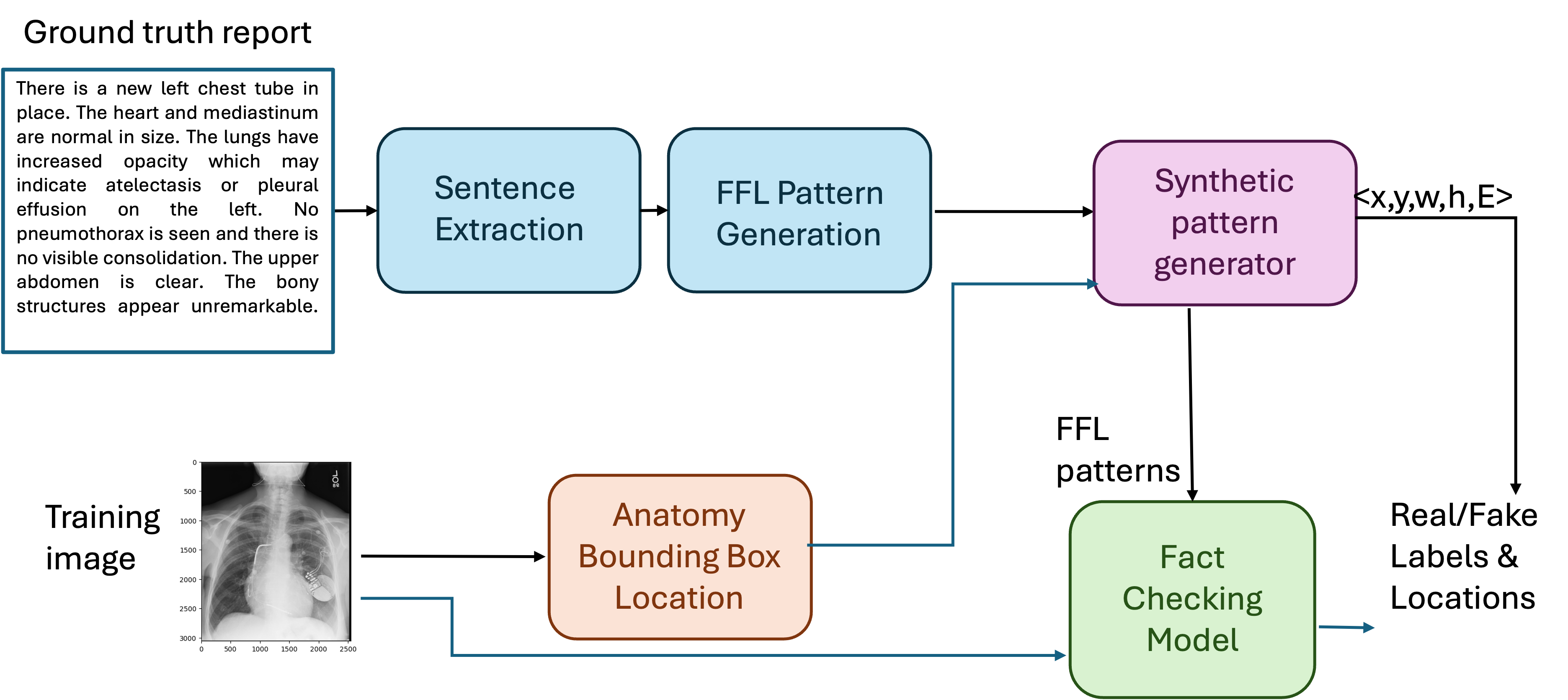}}
 \caption{Illustration of the FC model training using real and synthetic FFL patterns drawn from ground truth reports. }
  \label{fcmodeltrain}
\end{figure}

\begin{figure*}
\centering
  \centerline{\includegraphics[width=5.0in]{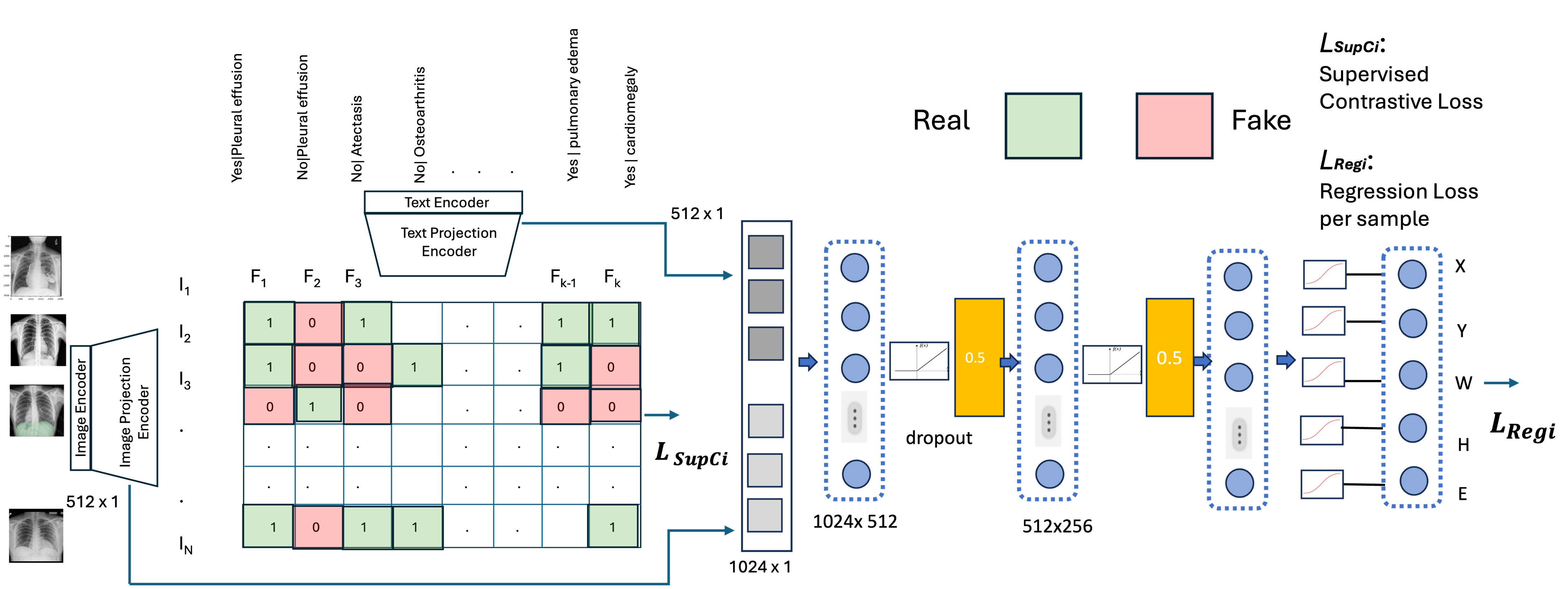}}
  \caption{Illustration of the architecture of our FC model consisting of a contrastive encoder and regression network. The real samples are taken as positive and the fake labels as negative in the contrastive formulation. The loss functions for the encoder and regressor are also shown in the figure. }
  \label{fcarchitecture}
\end{figure*}

The overall workflow for training our FC model is illustrated in Figure~\ref{fcmodeltrain} where the dataset of synthetic and real FL pairs along with their images are used to train our fact-checking model.  Given a mini batch B of training dataset of images $I=\{I_{i}\}$, and finding-location pairs $\{FL_{i}\}=\{FL_{iReal}\cup FL_{iFake}\}$, we learn a fact-checking model (FC-Model) that separates real pairs $(I_{i},FL_{iReal})$ from fake pairs $(I_{i},FL_{iFake})$. For this, we learn a suitable representation space in which images are brought close to their real FFL labels and separated from their fake labels using a contrastive encoder. The resulting representations of images and text are combined to learn the veracity of the finding and its location using a regression sub-network. The overall end-to-end architecture of the FC model is illustrated in Figure~\ref{fcarchitecture}. 


{\noindent\bf Multi-label cross-modal contrastive encoder:}

For building the encoder, we consider the finding labels of the FL-pairs only as $F_{iReal}$ and $F_{iFake}$ taken respectively from $FL_{iReal}$ and $FL_{iFake}$. Starting from a pre-trained CLIP model on chest X-rays \cite{Tiu2022,mlmi2023}, we train its image encoder (ViT-B/32 Transformer) and its encoder (masked self-attention Transformer) and their projection layers which are single linear layers (768x512 for image and 512x512 for text) by incorporating the pair-wise cosine similarity into a new multi-label supervised contrastive loss as given below. 

Let $z_{i}$ be the vision projection encoder output, and let $z_{f_{ij}}$ for each sample $D_{i}=(I_{i},F_{i})$ where $f_{ij}\in F_{i}=F_{iReal}\cup F_{iFake}$ are the real and fake labels per sample. Then we define a multi-label cross-modal supervised contrastive loss as:
\begin{equation}
\mathcal{L}_{SupC_{i}}=\frac{-1}{|F_{iReal}|} \sum_{f_{ij}\in F_{iReal}} log\frac{e^{s_if_{ij}/\tau}}{\sum_{a_{ik}\in F_{iFake}} e^{s_{ia_{ik}}/\tau}}
  \label{multilabelloss}
\end{equation}
\noindent where $s_{if_{ij}}=z_{i}\cdot z_{f_{ij}}$ is the pairwise cosine similarity between image and textual embedding vectors from the real findings $f_{ij}\in F_{iReal}$, and $s_{ia_{ik}}=z_{i}\cdot z_{a_{ik}}$ are from the cosine similarity with the fake findings where $a_{ik}\in F_{iFake}$. The overall loss is obtained by averaging across all the samples in the batch. Here $\tau$ is the temperature parameter.


{\noindent\bf Regression sub-network}


  \label{fcmodelinfer}
The output from the projection layers of image and text embeddings in the contrastive encoder are concatenated to form a new 1024-length feature vector which serves as the input to the regression subnetwork. The location information in our samples $D_{i}$ is now utilized to form the supervision label. Specifically, the input to the network is the vector $T_{ijReal}=[z_{i}|z_{f_{ij}}]$ formed from image $I_{i}$ and the real finding label $f_{ij}\in F_{iReal}$ or $T_{ijFake}=[z{i}|z_{a_{ik}}]$ where $a_{ik}\in F_{iFake}$ with the corresponding supervision label as the 5 regression parameters $(l_{ij},E_{j})$ where $l_{ij}=<x_{ij},y_{ij},w_{ij},h_{ij}>$ is the location of the finding $f_{ij}$ or $a_{ik}$ as the case may be and $E_{j}=1$ for the real finding $f_{ij}$ and 0 for the findings $a_{ik}$. The regression network consists of two linear layers, two drop out layers with RELU for intermediate layers and separate sigmoidal functions for producing the output regression vectors as shown in Figure~\ref{fcarchitecture}.

To reflect the dual attributes being optimized, namely, the location and the veracity of the finding, we form a combined loss function formed from mean square error loss, binary cross-entropy loss, L1-loss and generalized IOU loss. The L1 loss\cite{l1loss} and generalized IOU loss\cite{giou} have previously been used for regression\cite{chen23MedRPG}. However, in our case, since the negative findings have bounding box coordinates as $(0,0,0,0)$ it poses a problem in the generalized IOU computations when the prediction also gets close to the actual value. For this reason, and to ensure smooth convergence, we added the mean square penalty. Finally, for the veracity indicator variable $E$, we use the binary cross entropy loss. 

If we partition the output vector for each example into $Y=<Y_{1},Y_{2}>$ where $Y_{1}=<x,y,w,h>$ and $Y_{2}=E$, and the ground truth vector as $Y_{g}=<Y_{1g},Y_{2g}>$, we can then express the regression loss per sample as 
\begin{eqnarray}
\mathcal{L}_{Reg_{i}} =\underbrace{{\vert{Y_{1}-Y_{1g}\vert}}}_{\mathcal{L}_{1}(Y_{1}, Y_{1g})}+ \underbrace{\frac{\vert{Y_{1} \cap Y_{1g} \vert}}{\vert{Y_{1} \cup Y_{1g}}\vert} - \frac{\vert C_{Y_{1},Y_{1g}} \backslash {Y_{1} \cap Y_{1g}}\vert}{\vert{C_{Y_{1},Y_{1g}}}\vert} }_{\mathcal{L}_{\text{giou}}(Y_{1}, Y_{1g}) } \nonumber \\
+ \underbrace{{\vert{Y-Y_{g}}\vert}^{2}}_{\mathcal{L}_{\text{mse}}(Y, Y_{g})}  - \underbrace{[{Y_{2g}\textbf{log}(Y_{2})} + {(1-Y_{2g})\textbf{log}(1- Y_{2})}]}_{\mathcal{L}_{BCE}(Y_{2},Y_{2g})}
\label{regressionloss}
\end{eqnarray}
where $C_{Y_1,Y_{1g}}$ is the convex hull of the bounding boxes defined by $Y_{1}$ and $Y_{1g}$

{\noindent\bf FC Model Training:}

To train this network in an end-to-end fashion, the losses defined in Equations~\ref{multilabelloss} and ~\ref{regressionloss} were applied at the respective heads shown in Figure~\ref{fcarchitecture}. Starting from a CXR-pre-trained vision and text encoder\cite{Ramesh2022}, we trained the FC model for 100 epochs using the AdamW optimizer. The cosine annealing learning rate scheduler was used with the maximum learning rate of 1e-5 and 50 steps for warm up. An Nvidia A100 GPU with 40GB of memory was used with a batch size of 32.  The regression sub-network had a small number of parameters (657,413) in comparison to the pre-trained contrastive encoder ( 151,277,313 parameters). 


\begin{table*}

\centering
\begin{tabular}{lcccccc}
\hline
\multirow{2}{*}{\textbf{Dataset}} & \textbf{Patients} & \textbf{ Images} & \textbf{ Findings} & \textbf{Regions} & \textbf{Real/Synth Samples} \\
& \textbf{Train/Val/Test} &  &  & &
\\
\hline
ChestImagenome Silver\cite{Wu2021} & 44,133/6274/12,538 & 243,311 & 49 & 922,295 & 1,616,852/27,047,054\\
MS-CXR\cite{mscxr} & 478/54/114 & 925 & 8 & 2,254 & 2,247/24,338 \\
ChestXray8\cite{ChestXay8} & 457/51/109 & 880 & 8 & 1,571 & 1,571/10,137 \\
VinDr-CXR Train\cite{vindrcxr} & 9,450/1,050/2,250 & 15,000 & 23 & 69,052 & 47,973/132,632  \\
{\bf Chest ImaGenome Gold}\cite{Wu2021} & 390 & 439 & 35 & 5,477 & 4,063/23,463\\
\hline
\end{tabular}
\caption{Details of datasets used in experiments.}\label{datasets}
\end{table*}

{\noindent\bf Inference with FC model}
\label{fcinference}

The same pre-processing workflow shown for model training in Figure~\ref{fcmodeltrain} is repeated on the automated reports and the given image, to derive bounding box locations of anatomical regions of reported findings as shown in the inset of  Figure~\ref{radiologyworkflow}. The textual mention of the finding location in the FFL pattern is used to index the geometric location of the corresponding anatomical region as the indicated location as shown in the inset.  Next, the FC model is then applied to predict the real/fake label as well as the predicted location of the finding. Using the predicted locations, and the $E_{j}$ values per finding, an explainable visualization can be easily created through GUI tools as shown in Figure~\ref{radiologyworkflow}. 

{\noindent\bf Assessing automated report quality}
\label{fcquality}

We can use the FC model as a surrogate for ground truth to now assess the quality of the report in a quantitative way.  Specifically, let the FL-pairs extracted from the automated report $A$ for an image $I$ be denoted by $FL_{A}=(F_{A}=\{F_{Aj}\},L_{A}=\{L_{Aj}\})$ where $F_{A}$ are the FFL patterns found in the automated report, and $L_{A}$ are the indicated locations of $F_{A}$ in $I$. Let the predictions from the FC model be denoted by $(E_{p}=\{E_{j}\},L_{p}=\{L_{pj}\})$ where $E_{j}$ corresponds to the predicted label for the FFL pattern label $F_{Aj}\in F_{A}$  and $|F_{A}|=|E_{p}|$ and $L_{pj}$ corresponds to the indicated location $L_{Aj}\in L_{A}$ and $|L_{A}|=|L_{p}|$. Then the assessment score using the FC model can be summarized as FC-score(A,P) reflecting fraction of labels predicted as real and their relative overlap between the predicted and indicated locations as:
\begin{equation}
\mbox{FC-score(A,P)}=\frac{1}{2}(\frac{|E_{j}=1|}{\sum_{E_{j}\in E_{p}}E_{j}}+\frac{1}{|L_{p}|}{\sum_{j}
\frac{\vert{L_{Aj} \cap L_{pj} \vert}}{2\vert{L_{Aj} \cup L_{pj}}\vert}} )
\label{fcscore}
\end{equation}

{\noindent \bf Report correction:} To correct the automated reports, the findings that were flagged as an error ($E_{j}=0$) and their corresponding sentences are isolated. Since FFL pattern extraction algorithm records the location of the words in the sentence that mapped to the FFL pattern\cite{syeda-mahmood2020}, we can easily remove those words from the sentence. This leaves a fragmented sentence which is then given as an input to a large language model (Llama3.2) using a prompt 'Please make this a well-formed sentence'. The sentence returned by the LLM is then used to replace the original sentence in the automated report. Table~\ref{reportcorrection} shows examples of report sentences corrected through the LLM in this manner. More such examples are available in the supplementary material. 
\begin{table*}
\begin{small}
  \centering
  \begin{tabular}{c|c|c}
    \toprule
   Original Sentence & Error Finding & LLM-Corrected Sentence \\
    \midrule
    Left-sided pleural effusion found  & yes$|$pleural effusion$|$left lung & An abnormality was found,\\
    and the right atelectasis still remains.& &and the right atelectasis still remains.\\
    \hline
    The chest x ray image shows no focal  &   no$|$ pneumothorax & The chest X-ray image shows no focal\\ 
    consolidation, pulmonary edema,  & &     consolidation,  pulmonary edema, pleural effusion, \\
   pleural effusion or pneumothorax & & or other significant abnormalities.\\
    \bottomrule
  \end{tabular}
    \end{small}
  \caption{Illustration of LLM-based report correction. The first column shows a sample original sentence in which the finding classified as fake by the FC model is shown in the second column. The corrected sentence by LLM can add filler words as seen from the last column.}
  \label{reportcorrection}

\end{table*}
\begin{table*}
\centering
\begin{tabular}{c|c|c|c|c|c|c|c|c}
\hline
\multirow{2}{*}{\textbf{Method}} &  \multicolumn{2}{|c|}{\textbf{ChestImaGenome}} &\multicolumn{2}{|c|}{\textbf{MS-CXR}} &\multicolumn{2}{|c|}{\textbf{ChestX-ray8}} &\multicolumn{2}{|c}{\textbf{VinDR-CXR}}\\
& Accuracy & MIOU & Accuracy & MIOU & Accuracy & MIOU & Accuracy & MIOU \\
\hline
{\bf FCRegComb.}& {\bf 0.92} &  {\bf 0.54} & {\bf 0.94} & {\bf 0.53} & {\bf 0.92} & {\bf 0.57} & {\bf 0.90} & {\bf 0.49}\\
FCRegBCE & 0.88 &  0.49 & 0.92 & 0.46 & 0.90 & 0.53 & 0.88 & 0.45\\
FCRegDual & 0.87 &  0.51 & 0.89 & 0.49 & 0.87 & 0.51 & 0.86 & 0.47\\
FCRegSep  & 0.89 &  0.38 & 0.89 & 0.39 & {\bf 0.92} & 0.42 & 0.89 & 0.37\\
\hline
Med-RPG & -  & 0.23 & - & 0.32 & - & 0.28 & - & 0.38\\
Real/Fake Model & 0.84 & - & 0.78 & - & 0.81 & - & 0.83 & -\\
\hline
\end{tabular}
\caption{This table illustrates multiple aspects of the FC model evaluation. The FC model performance under different ablation architecture configurations across multiple datasets are rows in the first 4 rows. The last two rows comparison of our FC model's phrasal grounding and real/fake classification performance against SOTA methods. }\label{comparison}
\end{table*}
\begin{figure}
\centering
  \centerline{\includegraphics[width=3.0in]{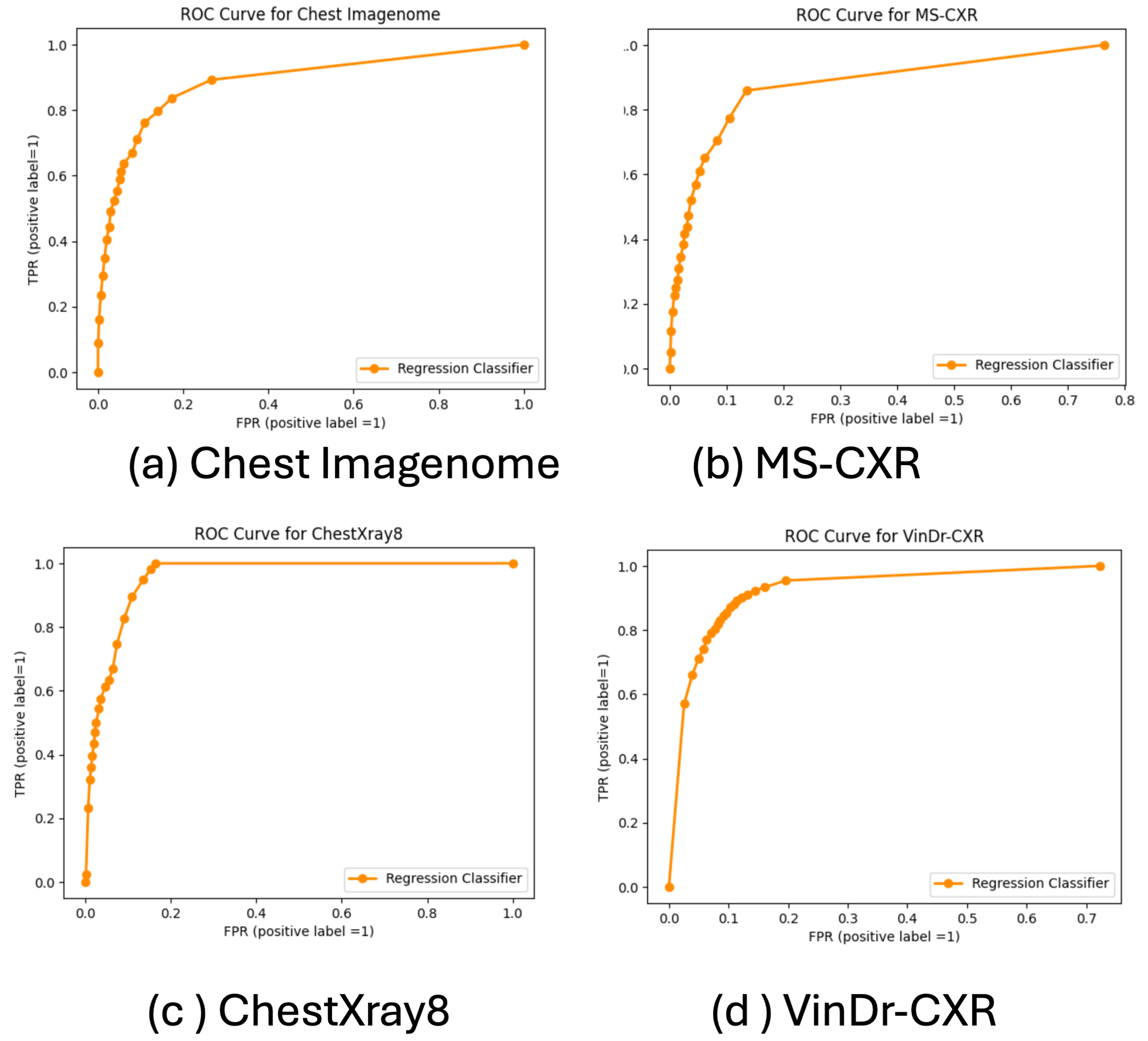}}
  \caption{Illustration of FC model accuracy in real/fake classification across the test splits of multiple datasets.(a) Chest ImaGenome Gold dataset (b) MS-CXR, (c) ChestXray8 from NIH, and (d) VinXrDR.}
  \label{roccurves}
\end{figure}
\section{Results}
We now describe several experiments conducted to evaluate the accuracy and efficacy of the model in error detection and correction of automated reports.

\noindent{\bf 3.1 Datasets:}
We selected 5 annotated chest X-ray datasets which had both location and finding annotations for our model evaluations as shown in Table~\ref{datasets}. 
Of these, the training partition of the ChestImagenome silver dataset was the largest and was used for training the FC model.  In addition, we selected the ChestImagenome gold dataset for our report quality evaluation as it had a complete set of ground truth reports, verified findings and their locations\cite{Wu2021}. Both silver and gold datasets are derived from MIMIC-CXR\cite{Journal:Johnson:arXiv2019}. The findings in the VinXrDR dataset were chosen as the reference as they had the most overlap among the datasets. More details are available in the supplementary material. 
{\noindent\bf Automated report generators:} To show the general applicability of our FC model to report generators, we selected several SOTA report generators whose GitHub code was freely available. These included 
RGRG\cite{rgrg}, XrayGPT\cite{xraygpt},
R2GenGPT\cite{R2GenGPT},
CV2DistillGPT2\cite{CV2DistillGPT2}, and an in-house hospital implementation of GPT-4 (GPT4-inhouse). Reports were collected from all the report generators on the ChestImagenome Gold dataset of 439 images using a common prompt of {\em 'For the input chest radiograph, please create a report based on radiographic findings.'} The reports derived from fact-checking and report correction were also recorded for the same images and compared to the ground truth reports for report quality evaluation. 



\noindent{\bf Explainable fact-checking}:
We first illustrate explainable fact-checking achieved by our model in Figure~\ref{exampleresults}. This shows several cases of errors flagged by our fact-checking model in the automated reports generated by XrayGPT\cite{xraygpt}.   In the visual explanation (top row), the indicated location computed from the reported findings as described in Section~\ref{fcinference} are shown drawn in yellow on the images of Figure~\ref{exampleresults} while the predictions from the model are shown in green.
The table in this figures shows all relevant details.  In each case, it can be seen that the FC model correctly flagged the errors and its predicted locations have a larger overlap with the ground truth location (shown in red) in comparison to the indicated locations from the automated reports.  

\noindent{\bf Real/Fake classification performance:}
We evaluated the accuracy of real/fake label prediction using the test partitions of the datasets shown in Table~\ref{datasets}.  The model consistently yielded an accuracy over 88\% for real/fake classification, as shown in Table~\ref{comparison} and by the ROC curves in Figure~\ref{roccurves} (see additional loss curves in supplemental materials). By using 10 fold cross-validation in the generation of the (70-10-20) splits for the datasets, the average accuracy of the test sets lay in the range 0.88 ± 0.12.

\medskip
\noindent{\bf Anatomical grounding performance:} 
We evaluated the anatomical grounding performance using mean IOU with the ground truth bounding boxes per sample. For each dataset, the mean IOU ranged from 0.49-0.57 as shown in Table~\ref{comparison} (rows 1-4), across various model architecture choices. 

\noindent{\bf 3.5 Comparison to related methods:} Since there was no prior work on explainable fact-checking with phrasal grounding, we compared separately to the nearest methods of only phrase grounding of chest x-ray findings, namely, MED-RPG\cite{chen23MedRPG}, and to fact-checking with only real/fake classification, called the Real/Fake Model\cite{mlmi2023}.  The results are shown in Table~\ref{comparison} in the last two rows recording the relevant numbers available for a classifier or regressor respectively. In comparison to pure phrase grounding or real/fake classification only, our method predicts both veracity and location of findings and outperforms these methods across all the datasets tested. 

\begin{table*}
\begin{tabular}{c|c|c|c|c|c|c|c|c}
\hline
\multirow{2}{*}{\textbf{Report Generator}} &  \multicolumn{2}{|c|}{\textbf{ChestImaGenome}} &\multicolumn{2}{|c|}{\textbf{MS-CXR}} &\multicolumn{2}{|c|}{\textbf{ChestX-ray8}} &\multicolumn{2}{|c}{\textbf{VinDR-CXR}}\\
& FC-Score & FC-Score& FC-Score & FC-Score & FC-Score &  FC-Score &  FC-Score &  FC-Score \\
& (A,P) & (A,G)& (A,P) & (A,G) & (A,P) & (A,G) &  (A,P) & (A,G) \\
\hline
RGRG\cite{rgrg}  & 0.459 &  0.463 & 0.671 & 0.692 & 0.695 & 0.702 & 0.451 & 0.463\\
XrayGPT\cite{xraygpt} & 0.378 & 0.374 & 0.612 & 0.609 & 0.623 & 0.645 & 0.382 & 0.391\\
GPT4-inhouse & 0.342 &  0.347 & 0.567 & 0.574 & 0.601 & 0.592 & 0.364 & 0.370\\
R2GenGPT\cite{R2GenGPT} & 0.413 &  0.415 & 0.623 & 0.626 & 0.654 & 0.667 & 0.419 & 0.421\\
CV2DistillGPT2\cite{CV2DistillGPT2} & 0.424 & 0.427 & 0.561 & 0.567 & 0.573 & 0.580 & 0.412 & 0.4\\
CheXRepair\cite{Ramesh2022} &  0.256 & 0.267 & 0.534 & 0.539 & 0.561 & 0.568 & 0.291 & 0.286\\
\hline
\end{tabular}
\caption{Illustrating the effectiveness of FC score-based assessment  as a surrogate ground truth during inference by comparing to ground truth-based assessment. The FFL patterns and their locations derived from different automated report generation methods indicated in Column 1 were used for the computation of the FC-score in both cases. (A,P) denotes the FC-score computed from automated and predicted findings from FC model. (A,G) denotes the FC score computed by matching the FFL patterns of automated reports and their locations with the ground truth. }\label{assessmentscore}
\end{table*}
\begin{table*}
\centering
\begin{tabular}{l|l|l|l|l|l|l|l|l}
\hline
{\bf Report} & {\bf \# Reports} & {\bf BLEU} & {\bf BLEU} & {\bf CheXbert} & {\bf CheXbert} & {\bf RadGraph} & {\bf RadGraph} & {\bf Avg.}\\
{\bf Generator} & {\bf \# Reports} & {\bf (A,G)} & {\bf (C,G)} & {\bf  (A,G)} & {\bf (C,G)} & {\bf F1 (A,G)} & {\bf F1 (C,G)} & {\bf Improve.}\\
\hline
{\bf RGRG}\cite{rgrg} &{\bf 439} & {\bf 0.237} & {\bf 0.315} & {\bf 0.329} & {\bf 0.444} & {\bf 0.529} &{\bf 0.741} & 35.6\%\\  
XrayGPT\cite{xraygpt} & 439 & 0.145 & 0.191& 0.256 & 0.345 & 0.390 & 0.565 & 37.5\%\\
GPT4 & 439 & 0.106 &0.148 & 0.087 & 0.131 & 0.434 & 0.607 & 43.3\%\\
R2GenGPT\cite{R2GenGPT} & 439 & 0.213 &  0.304	 & 0.353 & 0.461 & 0.237 & 0.357 & 41.5\%\\
CV2DistillGPT2\cite{CV2DistillGPT2} & 439 &  0.221 & 0.289 & 0.324 & 0.437 & 0.392 & 0.568 & {\bf 45.1\%}\\
CheXRepair\cite{Ramesh2022} & 439 &  0.069 & 0.093 & 0.134 & 0.194 & 0.256 & 0.363& 40.3\%\\
\hline
\end{tabular}
\caption{Illustration of improvement in report quality due to fact-checking and report correction on the ChestImagenome Gold dataset of ground truth report. (A,G) denotes report comparison of automated to ground truth report. (C,G) denotes report comparison of corrected report to ground truth report. Popular report evaluation scores based on lexical (BLEU), semantic (CheXbert), and clinical accuracy (Radgraph F1) are used for the analysis. All automated reports show improvement through the correction process although the largest improvement is seen using the Radgraph F1 score as it measures the clinical accuracy. }\label{scorecomparison}
\end{table*}
 

\noindent{\bf 3.4 Ablation studies:} 
Ablation studies were conducted to study the role of feature extraction and regression, the role of the loss function, its optimization, and the effect of end-to-end training. Specifically, we explore 4 architectures, namely,  (a) end-to-end network with trainable supervised contrastive loss encoder and regressor as depicted in Figure~\ref{fcarchitecture} (FCRegComb), (b) Replacing the loss with binary cross-entropy loss (BCE) for the encoder (FCRegBCE), (b) Using a generic pre-built CLIP encoder with regressor (FCRegSep) and (d) using a dual head regressor with separate loss functions for regression and classification (FCRegDual).   The results of real/fake classification and phrasal grounding is shown in Table~\ref{comparison}. As can be seen, combining the contrastive encoder with the regressor in an end-to-end fashion gave the best performance, justifying our choice of the model architecture.

{\noindent\bf FC model assessment evaluation:}
To evaluate the effectiveness of the FC-score in assessing automated report quality, we generated a similar FC-score from the ground truth. Specifically, let the FL pairs $FL_{G}=(F_{G},L_{G})$ be the FFL patterns and their locations flagged from ground truth in the datasets shown in Table~\ref{datasets}.  Since the FC model does not detect missed findings, we restrict $F_{G}$ to those that match findings in $F_{A}$ from the automated reports. The corresponding FC-score (A,G) between ground truth report and automated report can then be used as the benchmark to compare with FC-Score(A,P). The results of these are summarized in Table~\ref{assessmentscore} averaged across all images in the test partitions of the datasets indicated in Table~\ref{datasets} and using all the automated report generators against these images.  As can be seen, the FC-score using our FC model has good concordance with the corresponding FC-score from the ground truth pointing to the promise of the FC-score as a surrogate ground truth during inference in clinical workflows. 

{\noindent\bf Report quality improvement evaluation:}
In the last set of experiments, we evaluated the overall utility of our approach by recording the relative improvement in the quality of the corrected report in comparison to the original automated report. For this experiment, we used the ground truth report as a common reference. Since now the reports are composed of full-fledged sentences, any of the existing report evaluation scores can be utilized including lexical, semantic or clinical accuracy scores\cite{bleuscore,bertscore,radgraph}. The results of applying these pair-wise is shown in Table~\ref{scorecomparison}.  From this, we observe that on the average improvement in quality is around 40\% by employing the FC model and all automated reports were improved by the use of our FC model. 
\vspace{-0.07in}
\section{Discussion \& Conclusions}
In this paper, we have presented a new fact-checking approach that detects errors in the identity and location of findings. It also corrects the reports to lead to improved quality in the resulting automated reports.
From the results we see that this is possible even when the FC-model itself is not perfect in its prediction accuracy. Furthermore, no customization was needed for the FC model when a different choice of the report generator is available. Future work will address the current limitations of the model in handling omitted findings from reports. Overall, our paper showed that by carefully constructing synthetic datasets designed to elicit errors, we can develop discriminative models to correct the output of generative models at inference time, a result that may have significance beyond the domain of chest X-rays. 

{\small
\bibliographystyle{ieee_fullname}
\bibliography{egbib}
}

\end{document}